\documentclass[lettersize,journal]{IEEEtran}
\usepackage{amsmath,amsfonts}
\usepackage{algorithmic}
\usepackage{algorithm}
\usepackage{array}
\usepackage[caption=false,font=normalsize,labelfont=sf,textfont=sf]{subfig}
\usepackage{textcomp}
\usepackage{stfloats}
\usepackage{url}
\usepackage{verbatim}
\usepackage{graphicx}
\usepackage{cite} 
\usepackage{graphicx}
\usepackage{verbatim}
\usepackage{amsmath}
\usepackage{amssymb}
\usepackage{ctable}
\usepackage{multirow}
\usepackage{url}
\usepackage{marvosym}
\usepackage{rotating}
\usepackage{color}
\usepackage{sidecap}
\usepackage{colortbl}
\usepackage{xcolor} 

\newtheorem{thm}{Theorem}

\definecolor{MyColor}{HTML}{00FFFF} 
\definecolor{MyColorLess}{HTML}{06FB06} 
\definecolor{MyColorMore1}{HTML}{F9A5F5} 
\definecolor{MyColorMore2}{HTML}{FE4CA7} 
\definecolor{MyColorMore3}{HTML}{FFBF00} 
\definecolor{MyColorMore4}{HTML}{FF7E00}

\begin{document}
\sloppy 

\title{On-the-fly Global Embeddings Using 
\\Random Projections 
for Extreme Multi-label Classification}

\author{Yashaswi Verma 
\thanks{This paper was produced by the IEEE Publication Technology Group. They are in Piscataway, NJ.}
\thanks{Manuscript received Month Date, Year; revised Month Date, Year.} 
\thanks{Yashaswi Verma is with the Department of Computer Science and Engineering, 
Indian Institute of Technology, Jodhpur, India. (e-mail: yashaswi@iitj.ac.in).}}

\markboth{Journal of \LaTeX\ Class Files,~Vol.~14, No.~8, August~2021}%
{Shell \MakeLowercase{\textit{et al.}}: A Sample Article Using IEEEtran.cls for IEEE Journals}

\IEEEpubid{0000--0000/00\$00.00~\copyright~2021 IEEE}

\maketitle

\begin{abstract}
The goal of eXtreme Multi-label Learning (XML) is to automatically annotate a given data point with the most relevant subset of labels from an extremely large vocabulary of labels (e.g., a million labels). Lately, many attempts have been made to address this problem that achieve reasonable performance on benchmark datasets. In this paper, rather than coming-up with an altogether new method, our objective is to present and validate a simple baseline for this task. Precisely, we investigate an on-the-fly global and structure preserving feature embedding technique using random projections whose learning phase is independent of training samples and label vocabulary. Further, we show how an ensemble of multiple such learners can be used to achieve further boost in prediction accuracy with only linear increase in training and prediction time. Experiments on three public XML benchmarks show that the proposed approach obtains competitive accuracy compared with many existing methods. Additionally, it also provides around 6572$\times$ speed-up ratio in terms of training time and around 14.7$\times$ reduction in model-size compared to the closest competitors on the largest publicly available dataset. 
\end{abstract}


\section{Introduction} 
\label{sec:Introduction} 
\IEEEPARstart{e}{Xtreme} Multi-label Learning (or XML) is the problem of 
learning a classification model that can automatically 
assign a subset of the most 
relevant labels to a data point from an extremely large 
vocabulary of labels. 
With a rapid increase in the digital content on the Web, such 
techniques can be useful in several 
applications such as indexing, tagging, ranking 
and recommendation, and thus their requirement is 
becoming more and more critical. 
{\it E.g.}, Wikipedia contains more than a million labels 
and one might be interested in learning a model using 
the Wikipedia articles and corresponding labels, so that it 
can be used to automatically annotate a new article 
with a subset of the most relevant labels. 
Another example can be to display a subset of advertisements 
to online users based on their browsing history. 
It is important to note that multi-label classification 
is different from multi-class classification that aims 
at assigning a {\em single} label/class to a data point. 

While XML has several applications, it is a 
challenging problem as it deals with a very 
large number (hundreds of thousands, or even millions) 
of data points and features, and a practically 
unimaginable size of 
the output space (exponential in terms of vocabulary size). 
As a result, recently 
this has been approached using several interesting 
techniques such as 
\cite{sleec,leml,pfastrexml,fastxml,dismec,pdsparse,ppdsparse,annexml,xmlcnn,deepxml,parabel,bonsai,xtransformer,attentionxml,astec,aplcxlnet}, 
most of which try to capture the relationships 
between features and labels. 
Moreover, these also attempt to 
address the scalability aspect that is particularly critical 
in the XML task, and is usually achieved by 
making an extensive use of computational resources. 
While many of these techniques have shown good 
performance, one thing that is missing in the XML literature 
is comparison with a conceptually simple and computationally 
light technique that can justify 
the need for complex models and resource-intensive training. 
In this paper, we are interested in investigating a simple 
approach for XML that can fill this gap and 
thus serve as a baseline for this task. 
%

With this objective, we present On-the-fly Global 
Embeddings for Extreme Classification 
(OGEEC, pronounced as ``o-geek''), 
an XML classifier than can quickly generate a global 
and structure preserving feature embedding, and easily scale 
to very large 
datasets using limited computational resources. 
We start by posing XML as a retrieval task where given a 
new data point, we retrieve its k-nearest neighbours 
from the training set, and then 
perform a weighted propagation of 
the labels from the nearest points to the input point 
based on their degree of similarity. 
While this idea sounds simple and straightforward, applying 
this directly in XML 
is practically difficult, 
because computing 
the nearest neighbours from a large training set 
in a very high-dimensional feature space 
(both are of the order $10^{5}-10^{6}$) 
is computationally prohibitive~\cite{leml,sleec,annexml}.   
To address this, we adopt a global and linear feature 
embedding technique 
that performs an inherently non-linear projection of 
a high-dimensional data into a lower dimensional 
space. 
This computationally linear yet 
inherently non-linear projection can be computed in a 
couple of seconds (hence we call it {\it on-the-fly}), and 
also preserves pairwise 
distances among data points within a certain error 
bound, thus allowing 
efficient computation of the nearest neighbours in 
a low-dimensional (of the order $10^{2}$) latent 
embedding space. 
Precisely, this embedding is inspired from the strong 
theoretical properties of the JL-Lemma~\cite{johnson84}, 
and is quite simple: 
compute a matrix whose entries are independently sampled 
from the Normal distribution, 
and use it for feature projection/embedding into a 
lower dimensional space. 
It is interesting to note here that while this embedding is 
obtained without using the training data, it is still 
capable of preserving the structure of the data, 
thanks to the large sample-size in XML. 
In other words, while 
an extremely large amount of data in XML becomes 
a challenge for other learning techniques, it is 
conducive for applying the JL-Lemma. 
Further, to safeguard against the randomness that is implicit in the 
process of generating an embedding, we generate an ensemble of 
embeddings (or learners) by doing multiple samplings from 
the Normal distribution. 
Our empirical analyses demonstrate that using such ensembles 
leads to not only stable solutions 
with only linear increase in training and prediction time 
but also significant increase 
in prediction accuracies 
which are 
quite competitive (and many a times even better) 
compared to the state-of-the-art XML methods. 

To summarize, the contributions of this paper are: 
(1) We present OGEEC classifier for XML which is simple, 
accurate, and thus can be used 
as a baseline for comparing other XML algorithms. 
To demonstrate the little resource requirements 
of OGEEC, 
we conduct all the experiments (both 
training as well as testing) on a single 
CPU core of an eight-core desktop 
(Intel i$7$-$7700$ $3.60$GHz $\times$ $8$ processor 
and $15.6$ GB RAM). 
(2) We supplement the study with extensive experimental analyses 
and comparisons with several state-of-the-art XML methods 
on three large-scale and benchmark XML datasets 
(Delicious-200K, Amazon-670K and Amazon-3M). 


\section{Related Work} 
\label{sec:RelatedWork}

The XML problem has been approached by both 
non-deep learning based as well as deep learning based 
approaches. Among the non-deep learning based approaches, 
the three broad sub-categories include 
feature embedding-based approaches~\cite{sleec,leml,reml,annexml}, 
linear (one-vs-all) classification-based 
approaches~\cite{dismec,pdsparse,ppdsparse,proxml}, and 
tree-based approaches 
\cite{lpsr,fastxml,pfastrexml,craftml,bonsai,slice,parabel}. 
Among the deep learning based approaches, the existing 
approaches can be categorized on the basis of whether 
they learn global features~\cite{xmlcnn,deepxml,astec,aplcxlnet} 
or local/attention-based features~\cite{attentionxml,xtransformer}. 

The embedding based approaches focus on reducing the effective 
number of features, labels or both by projecting them into 
a lower dimensional space. Among the initial approaches 
such as LPSR-NB~\cite{lpsr}, LEML~\cite{leml} and REML~\cite{reml}, 
while LPSR-NB uses label hierarchies to iterative partition 
the label space, LEML and REML learn to 
project label-matrix into a low-rank structure and 
perform a projection back into the original label space 
during prediction. 
Recently, SLEEC~\cite{sleec} 
and AnnexML~\cite{annexml} have been considered as the 
representative methods in this category. 
These consist of three steps: clustering the 
samples, learning a non-linear 
low-dimensional feature embedding for each cluster, and 
kNN-based classification. During 
training, the low-dimensional embedding is learned such that it 
preserves pairwise distances between closest label vectors, thus 
capturing label correlations. The clustering of training samples 
helps in speeding-up the testing process, as the neighbours 
of a test sample are computed only from the group to which it 
belongs. The difference between them is that 
SLEEC aims at preserving pairwise distances wheres AnnexML 
uses a graph embedding method such that 
the k-nearest neighbour graph of the samples is preserved. 
Since clustering high-dimensional features is usually unstable, 
both SLEEC as well as AnnexML further use an ensemble of such 
learners by using different clusterings, and the predictions 
from all the learners in an ensemble are averaged to get the 
final prediction.

The linear classification-based approaches such 
as~\cite{dismec,proxml,pdsparse,ppdsparse} learn a 
linear classifier per label, with the difference being in the 
form of loss function and constraints imposed on the 
classifier. DiSMEC~\cite{dismec} learns max-margin 
classifiers using the conventional hinge-loss and 
L2-norm regularization. To scale this to large vocabularies, it 
uses few hundreds of CPU cores in parallel. Before testing, the 
dimensions in each classifier which have negligible magnitude 
are explicitly omitted for speed-up. To address this, 
L1-norm regularization is adopted in 
ProXML~\cite{proxml} that helps in learning sparse classifiers, 
which is also shown to be helpful in boosting the prediction 
accuracy of rare labels. 
In PD-Sparse~\cite{pdsparse} and PPD-Sparse~\cite{ppdsparse},  
negative-sampling through a sparsity preserving optimization 
is used to reduce the training-time, and heuristic-based 
feature sampling is performed to reduce the prediction time. 
In general, while these approaches achieve high prediction 
accuracies, their training and prediction times increase 
significantly and it may take 
several weeks to train these models on large vocabulary datasets. 

The third direction is based on tree based 
approaches such as~\cite{plt,firstxml,lpsr,fastxml,pfastrexml,craftml,parabel,bonsai,slice} 
that perform hierarchical feature/label-based partitioning for 
fast training and prediction. However, due to the cascading 
effect, an error made at a top level propagates to lower levels. 
Among these, FastXML~\cite{fastxml}, PfastreXML~\cite{pfastrexml} 
aim at optimizing label ranking by recursively partitioning the feature space. 
However, since these learn a weak classifier at each node in a tree 
for fast traversal, this affects their prediction accuracy. 
This is overcome in 
a recent hybrid approach method Parabel~\cite{parabel} 
that learns high-dimensional binary classifiers 
similar to~\cite{dismec} for label partitioning at each node 
and achieves promising results, though 
with increased complexity and memory requirements. 
Taking motivation from these methods, other methods 
such Slice~\cite{slice} and Bonsai~\cite{bonsai} 
try to address some of their limitations. 
E.g., in Slice~\cite{slice}, rather than learning a 
classfier for label partitioning using all the labels, 
only the most confusing labels are used that are 
identified using an efficient negative 
sampling technique. 
Similarly, in Bonsai~\cite{bonsai}, rather than learning 
a deep but narrow tree, a shallow but relatively wider tree 
is learned by clustering the label representations using 
the k-means algorithm.

While non-deep learning based methods discussed above 
have rely on bag-of-words features, deep learning 
based methods use 
raw input to learn dense representations using multiple 
non-linear transformations that can 
effectively capture semantic as well as contextual 
information. 
One of the first such methods for XML was 
XML-CNN~\cite{xmlcnn} which uses 1-D CNN along both 
word embedding dimension and sequence length. The dense 
embedding learned by XML-CNN is also used in 
SLICE as discussed above that learns a tree-based model. 
Another approach DeepXML~\cite{deepxml} was the first 
to learn deep feature embedding in XML by 
using a label graph structure. 
A recent approach APLC-XLNet~\cite{aplcxlnet} fine-tunes 
a generalized autoregressive pre-trained model 
XLNet~\cite{xlnet} 
and forms clusters of labels based on 
label distribution to approximate the cross-entropy loss. 
Among the attention-based methods that learn local features, 
AttentionXML~\cite{attentionxml} uses 
BiLSTMs and label-aware attention, 
while X-Transformer~\cite{xtransformer} pre-trained using the 
BERT transformer model~\cite{bert}. 
In general, while these methods achieve good empirical results, 
their models are quite heavy with few hundreds/thousands of 
millions of learned parameters and are thus computationally demanding 
(with some of them requiring a cluster of GPUs for execution). 
In later parts of the paper, we will contrast our approach 
with these methods as well as analyze the pros and cons of each.



\section{The OGEEC Approach} 
\label{sec:BaselineMethods} 

In this section, we describe the proposed OGEEC approach. 
Let $\mathcal{D} = \{(\mathbf{x}_1,\mathbf{y}_1), 
\ldots, (\mathbf{x}_n,\mathbf{y}_n)\}$ denote the training set, 
where $\mathbf{x}_i \in \mathcal{X} \subseteq \mathbb{R}^d$ 
is a $d$-dimensional 
feature vector and 
$\mathbf{y}_i \in \mathcal{Y} \subseteq \{0,1\}^L$ is 
the corresponding binary label vector that denotes the 
labels assigned to $\mathbf{x}_i$, with $1$ denoting 
the presence and $0$ denoting the absence of the 
corresponding label. Let 
$\mathbf{X} = [\mathbf{x}_1; \ldots ; \mathbf{x}_n] 
\in \mathbb{R}^{d\times n}$ be the data matrix whose 
each column is a feature vector. 


One of the simplest learning based technique 
that can be employed to 
project high-dimensional features into a lower dimensional space 
is Principal Component Analysis (PCA)~\cite{hastie}. 
Though simple, the operations involved in PCA (such as 
eigenvalue decomposition) are quite expensive (in terms of both 
computation and memory), and can scale to 
only a few tens of thousands of 
feature dimensions~\cite{largePca11,largePca17} 
with heavy computation and memory 
requirements. 
Due to this, it is not feasible to use PCA for dimensionality 
reduction when the dimensionality of input features is in several 
hundreds of thousands or even millions as in XML. 
\footnote{Because of this, none of the existing XML methods has 
reported comparisons with PCA.}. 
%
On the other hand, Johnson and Lindenstrauss~\cite{johnson84} 
showed that the structure of high-dimensional data is well 
preserved in a lower dimensional space projected using 
random linear projections. As a result, random projections 
have been proven to be useful in a variety of applications 
such as dimensionality reduction, clustering~\cite{fern03}, 
denstiy estimation~\cite{dasgupta00}, etc. 
Below, we first give an overview of 
the JL-Lemma, and then present the proposed approach.

\subsection{Background: Johnson-Lindenstrauss Lemma (or JL-Lemma)} 
\label{sec:JLLemma} 

When we seek a dimensionality reduction where the goal is to 
preserve the structure of the data by preserving 
pairwise distances between the data points, we can 
make use of a projection matrix 
that is randomly initialized in a particular way. 
This is also known as the random projection method, and is studied 
under the JL-Lemma~\cite{johnson84}. 
Suppose we are initially given $n$ data points 
$\mathbf{x}_1, \ldots, \mathbf{x}_n \in \mathbb{R}^d$ 
in a $d$-dimensional space, and we are interested in 
projecting these points into a lower dimensional space 
and find $n$ points $\mathbf{a}_1,\ldots,\mathbf{a}_n \in \mathbb{R}^r$, 
where $r << d$, such that 
\begin{eqnarray} 
||\mathbf{a}_j|| \;\; \approx \;\; ||\mathbf{x}_j|| \;\;\; \forall j, \nonumber 
\\ 
||\mathbf{a}_i - \mathbf{a}_j|| \;\; \approx \;\; 
||\mathbf{x}_i - \mathbf{x}_j|| \;\;\; \forall i,j. \nonumber  
\end{eqnarray} 
where $||\mathbf{x}||$ denotes the Euclidean norm of the vector 
$\mathbf{x}$. Then, the JL-Lemma is given as follows: 
\begin{thm} 
Let $\mathbf{x}_1,\ldots,\mathbf{x}_n \in \mathbb{R}^d$ be 
arbitrary. Pick any $\epsilon = (0,1)$. Then for some 
$r = O(log(n)/\epsilon^2$) there exist points 
$\mathbf{a}_1,\ldots,\mathbf{a}_n \in \mathbf{R}^r$ such that 
\begin{eqnarray} 
(1-\epsilon)||\mathbf{x}_j|| \leq ||\mathbf{a}_j|| 
 \leq  (1+\epsilon)||\mathbf{x}_j||, \forall j \nonumber 
\\ 
\label{eq:jlErrorBounds}  
(1-\epsilon)||\mathbf{x}_i-\mathbf{x}_j|| 
 \leq  ||\mathbf{a}_i-\mathbf{a}_j|| 
 \leq  
(1+\epsilon)||\mathbf{x}_i-\mathbf{x}_j||, \forall i,j. 
\end{eqnarray} 
Moreover, in polynomial time we can compute a linear transformation 
$\mathbf{F}:\mathbb{R}^d \rightarrow \mathbb{R}^r$ such that, 
defining $\mathbf{a}_j = \mathbf{F}(\mathbf{x}_j)$, the 
inequalities in the above equation are satisfied with probability 
at least $1 - 2/n$. 
\end{thm} 
This implies that the linear transformation $\mathbf{F}$ 
can be used for projecting the initial data points into a 
lower dimensional space such that their pairwise distances 
are preserved within an error bound. 
In practice, the linear transformation $\mathbf{F}$ is 
a matrix whose entries are independent random 
variables sampled from a Normal 
distribution~\cite{jllemmaLink}. 
It is also important to note that the projected 
points $\{\mathbf{a}_1,\ldots,\mathbf{a}_n\}$ have no dependence 
on the dimensionality of the input samples (i.e., $d$), which 
implies that the original data could be in an arbitrarily high 
dimensional space, thus making it suitable for XML. 
Another important thing is 
that while PCA is useful only when the original data 
points 
inherently lie on a low dimensional manifold, 
this condition is not required by the JL-Lemma.

\begin{table}[t] 
\begin{center} 
\begin{tabular}{p{0.9\columnwidth}} 
\hline 
{\bf Algorithm 1:} Obtaining a single feature embedding 
matrix 
\\ 
\hline 
{\bf Require:} \\ 
(1) Training feature matrix: 
$\mathbf{X} = [\mathbf{x}_1; \ldots ; \mathbf{x}_n] 
\in \mathbb{R}^{d\times n}$ \\ 
(2) Embedding dimension: $r$ ($r \leq d$) 
\\ 
\hline 
{\bf Method:} \\ 
Step-1: Compute matrix: $\mathbf{F} = 
normrnd(0,1,r,d)$ 
$\;$ \\// in MATLAB 
\\ 
\hline 
{\bf Post-processing:} 
\\ 
Step-1: Normalize training features $\mathbf{X}$ 
using $\text{L}_\text{2}$-normalization \\ 
Step-2: $\mathbf{X} \leftarrow (\mathbf{X}^T\times\mathbf{F})^T$ \\ 
Step-3: Re-normalize training features $\mathbf{X}$ 
using $\text{L}_\text{2}$-normalization 
\\ 
\hline 
\end{tabular} 
\end{center} 
\end{table}

\subsection{Feature Embedding} 
\label{sec:ProposedEmbedding} 

Taking motivation from the JL-Lemma, 
we adopt a simple and easy to implement approach 
for generating a linear yet inherently non-linear 
and structure preserving projection matrix: 
we compute a matrix of random numbers generated from 
the Normal distribution (i.e., a Gaussian 
distribution with zero mean and unit variance, and 
use this to perform a linear 
projection of high ($d$-) dimensional input features into a 
lower ($r$-) dimensional latent 
embedding space, keeping $r<<d$. 
It is easy to note that obtaining this projection matrix 
does not involve any 
learning based on the given training data points. 
For completion, Algorithm~1 summarizes the steps of obtaining a 
single such global embedding. 
Later, in Section~\ref{sec:Datasets}, we will discuss 
the error bounds of the pairwise distances for 
different datasets and 
using different projection matrices. 


\begin{table*}[t] 
\caption{Multi-label datasets used for empirical analysis in this 
        paper (available at the Extreme Classification 
        Repository~\cite{xmlrepo}). 
        ASpL and ALpS denote the average number of samples 
        per label and average number of labels per sample respectively 
} 
\label{tab:DatasetStatistics} 
\begin{center} 
\begin{tabular}{ccccccc} 
\hline 
{Dataset} 
& {\# Training Samples} 
& {\# Test Samples} 
& {\# Features} 
& {\# Labels} 
& {ASpL} 
& {ALpS} 
\\ 
\hline 
 {Delicious-200K} & 
 {196,606} &  {100,095} & 
 {782,585} &  {205,443} & 
 {72.29} &  {75.54} 
\\ 
 {Amazon-670K} & 
 {490,449} &  {153,025} & 
 {135,909} &  {670,091} & 
 {3.99} &  {5.45} 
\\ 
 {Amazon-3M} & 
 {1,717,899} &  {742,507} & 
 {337,067} &  {2,812,281} & 
 {31.64} &  {36.17} 
\\ 
\hline 
\end{tabular} 
\end{center} 
\end{table*}

Next, in order to control the randomness involved in this 
process and achieve stable predictions, we obtain a set 
of multiple such global feature embeddings (learners) by using 
different random samplings from the Normal distribution. 
In particular, we do this by using different 
``seed'' values for initializing random matrices. 


\begin{table}[t] 
\begin{center} 
\begin{tabular}{p{0.9\columnwidth}} 
\hline 
{\bf Algorithm 2:} Label prediction using a single embedding 
\\ 
\hline 
{\bf Require:} \\ 
(1) Test point $\mathbf{x}_t \in \mathbb{R}^{d}$\\ 
(2) Number of nearest neighbours: $k$ \\ 
(3) Projected and normalized training features $\mathbf{X}$ \\ 
(4) Feature embedding matrix: $\mathbf{F}$ \\ 
\hline 
{\bf Pre-processing:} \\ 
Step-1: Normalize $\mathbf{x}_t$ using 
$\text{L}_\text{2}$-normalization \\ 
\hline 
{\bf Method:} \\ 
Step-1: $\mathbf{x}_t \leftarrow (\mathbf{x}_t^T \times \mathbf{F})^T$ \\ 
Step-2: Normalize $\mathbf{x}_t$ using 
$\text{L}_\text{2}$-normalization \\ 
Step-3: Compute the $k$-nearest neighbours of $\mathbf{x}_t$ from 
$\mathbf{X}$ using dot-product \\ 
Step-4: Propagate the labels from the neighbours by weighting them 
with the corresponding similarity scores\\ 
\hline 
\end{tabular} 
\end{center} 
\end{table}

\subsection{Label Prediction} 
\subsubsection{Using a single embedding} 

For label prediction, we 
use a weighted k-nearest neighbour based classifier to propagate 
labels to a new sample from its few nearest 
neighbours in the training 
set. For each label, we use Bernoulli models, considering 
either presence or absence of labels in the 
neighbourhood~\cite{mbrm}. 

Let $\mathbf{x}_t$ denote a test sample, $\mathcal{N}^k_t$ 
denote the set of its $k$ nearest neighbours from the 
training set 
(with similarity being computed using dot-product), and 
$y^w \in \{0,1\}$ denote the presence/absence of the label 
corresponding to index $w$ for a sample $\mathbf{x}$. 
Then, the label presence prediction for $\mathbf{x}_t$ is defined 
as a weighted sum over the training samples in $\mathcal{N}^k_t$: 

\begin{eqnarray} 
\label{eq:predProbPart1} 
p(y^w_t=1|\mathbf{x_t}) &\propto&  
\sum_{\mathbf{x}_i\in\mathcal{N}^k_t} \pi_{it} \;p(y^w_i=1|\mathbf{x}_i), 
\\ 
\label{eq:predProbPart2} 
p(y^w_i=1|\mathbf{x}_i) &=&  
\left\{ \begin{array}{rl} 
1 & \text{for}  \; y^w_i = 1, \\ 
0 & \text{otherwise,} 
\end{array}\right. 
\end{eqnarray} 

\noindent 
where $\pi_{it}$ denotes the importance of the 
training sample $\mathbf{x}_i$ 
in predicting the labels of the test sample $\mathbf{x}_t$, and 
is given by $\pi_{it} = max(\mathbf{x}_i^T\mathbf{x}_t,0)$. 
Since we assume the samples to be 
$\text{L}_\text{2}$-normalized, this is equivalent to 
computing the cosine similarity score between them. 
Using Eq.~\ref{eq:predProbPart1}, we get prediction 
scores for all the labels and pick the top-$K$ 
($K\in\{1,3,5\}$) for assignment and performance 
evaluation. It is important to 
note that while we do not explicitly model the dependencies 
among labels in the training data, these are implicitly 
exploited in our model. This is because the labels that 
co-occur in a given training sample get the same weight 
depending on the degree of similarity of that training sample 
with the test sample, thus 
implicitly capturing sample-to-sample, sample-to-label as well as 
label-to-label similarities. 
Algorithm~2 summarizes the steps of label prediction using 
a single learner. In case of multiple learners, 
we use 
Eq.~\ref{eq:predProbPart1} to get the prediction scores 
for all the labels using each learner individually. 
Then we average 
these prediction scores over all the learners in an 
ensemble, and pick the top few labels for assignment and 
performance evaluation. 

\subsection{Relation with deep learning based methods} 
\label{sec:RelationWithDeepLearning} 

In the last few years, deep learning based methods 
have become popular
for several tasks, particularly those that involve 
classification. These methods aim at learning 
characteristics/features that are
distinctive across categories. To do so, they 
learn a highly non-linear transformation 
function of the input data. 
The global embedding technique used in 
OGEEC shares a similar objective (specifically, with 
deep metric learning based methods) in the sense that it 
also involves an inherently non-linear and structure-preserving 
mapping of features. However, in contrast to deep 
learning based methods that rely on an extensive use of training 
data and compute resources, the global embedding 
used in OGEEC relies on the 
distributional properties of an extremely large data set in 
a high-dimensional space which are guaranteed by the JL-Lemma. 
This enables OGEEC to obtain a data-independent 
linear transformation that is inherently non-linear and 
structure-preserving. 
As we will discuss in Section~\ref{sec:ComparisonWithSota}, 
despite learning the transformation function 
without making use of training 
data, OGEEC still performs fairly well in comparison to 
deep learning based XML methods.

\section{Experiments} 
\label{sec:ExperimentalEvaluation} 

\subsection{Experimental Set-up} 

\subsubsection{Datasets and Analysis} 
\label{sec:Datasets} 

We use three large-scale XML datasets in our experiments: 
Delicious-200K, Amazon-670K 
and Amazon-3M, available at the Extreme Classification 
Repository~\cite{xmlrepo}. Among these, Amazon-670K 
and Amazon-3M are the top-two largest 
public XML datasets. 
We use the same training and test partitions as given in the 
repository, and do not 
use any additional meta-data. The statistics of these 
datasets are summarized in Table~\ref{tab:DatasetStatistics}. 
For all the datasets, we use the pre-computed bag-of-words 
features available at the above repository.

\begin{table}[h] 
\caption{Error bounds of the pairwise distances 
        (Eq.~\ref{eq:jlErrorBounds}) 
        for different datasets by considering 
        $r=200$ 
} 
\label{tab:ErrorBounds} 
\begin{center} 
\begin{tabular}{cccc} 
\hline 
{Dataset} 
& {${\epsilon}$} 
& {${1-\epsilon}$} 
& {${1+\epsilon}$} 
\\ 
\hline 
 {Delicious-200K} 
&  {0.1627} &  {0.8373} &  {1.1627} 
\\ 
 {Amazon-670K} 
&  {0.1687} &  {0.8313} &  {1.1687} 
\\ 
 {Amazon-3M} 
&  {0.1766} &  {0.8234} &  {1.1766} 
\\ 
\hline 
\end{tabular} 
\end{center} 
\end{table}

\begin{table}[h] 
\caption{Error bounds of the pairwise distances for the 
        Delicious-200K dataset by varying the dimensionality 
        of the projection space in $\{50,100,\ldots,400\}$ 
} 
\label{tab:VaryErrorBounds} 
\begin{center} 
\begin{tabular}{ccccccccc} 
\hline 
 {Output Ftr. Dimension ($r$)} 
& {${\epsilon}$} 
& {${1-\epsilon}$} 
& {${1+\epsilon}$} 
\\ 
\hline 
{50} & {0.3254} 
& {0.6746} & {1.3254} 
\\ 
{100} & {0.2301} 
& {0.7699} & {1.2301} 
\\ 
{150} & {0.1879} 
& {0.8121} & {1.1879} 
\\ 
{200} & {0.1627} 
& {0.8373} & {1.1627} 
\\ 
{250} & {0.1455} 
& {0.8545} & {1.1455} 
\\ 
{300} & {0.1328} 
& {0.8672} & {1.1328} 
\\ 
{350} & {0.1230} 
& {0.8770} & {1.1230} 
\\ 
{400} & {0.1150} 
& {0.8850} & {1.1150} 
\\ 
\hline 
\end{tabular} 
\end{center} 
\end{table}

\begin{table*}[ht] 
\caption{Performance of OGEEC in terms of 
        Precision@$K$, nDCG@$K$, propensity-scored 
        Precision@$K$, and propensity-scored nDCG@$K$ ($K$=1,3,5). 
        In case of ``One learner'', the accuracies 
        are averaged over five individual learners. 
        In case of ``Five learners'', the accuracies are 
        calculated after fusing the predictions obtained using 
        five individual learners as described in 
        Section~\ref{sec:ProposedEmbedding} 
} 
\label{tab:ResultsNoProp} 
\begin{center} 
\begin{tabular}{cc|cccc|cccc} 
\multicolumn{1}{c}{}
& \multicolumn{1}{c}{}
& \multicolumn{4}{c}{ {\bf One learner}} 
& \multicolumn{4}{c}{ {\bf Five learners}} 
\\ 
\hline 
{{Dataset}} 
& { {Metric$@K$}} 
&  {Prec.} &  {nDCG} 
&  {PS-Prec.} &  {PS-nDCG} 
&  {Prec.} &  {nDCG} 
&  {PS-Prec.} &  {PS-nDCG} 
\\ 
\hline 
\multirow{3}{*}{{Delicious-200K}} 
& {@1} 
&  {36.89$\pm$0.11} &  {36.89$\pm$0.11} 
&  {5.83$\pm$0.02} &  {5.83$\pm$0.02} 
&  {40.54} &  {40.54} 
&  {6.37} &  {6.37} 
\\ 
& {@3} 
&  {30.86$\pm$0.04} &  {32.28$\pm$0.05} 
&  {6.26$\pm$0.01} &  {6.14$\pm$0.01} 
&  {34.25} &  {35.74} 
&  {6.91} &  {6.76} 
\\ 
& {@5} 
&  {27.79$\pm$0.02} &  {29.91$\pm$0.03} 
&  {6.62$\pm$0.01} &  {6.39$\pm$0.01} 
&  {30.97} &  {33.22} 
&  {7.33} &  {7.05} 
\\ 
\hline 
\multirow{3}{*}{{Amazon-670K}} 
& {@1} 
&  {35.21$\pm$0.06} &  {35.21$\pm$0.06} 
&  {21.84$\pm$0.06} &  {21.84$\pm$0.06} 
&  {37.45} &  {37.45} 
&  {23.05} &  {23.05} 
\\ 
& {@3} 
&  {31.98$\pm$0.02} &  {33.77$\pm$0.02} 
&  {24.89$\pm$0.04} &  {24.09$\pm$0.03} 
&  {33.67} &  {35.64} 
&  {26.19} &  {25.37} 
\\ 
& {@5} 
&  {29.73$\pm$0.03} &  {33.00$\pm$0.03} 
&  {28.05$\pm$0.03} &  {26.21$\pm$0.03} 
&  {31.12} &  {34.68} 
&  {29.42} &  {27.54} 
\\ 
\hline 
\multirow{3}{*}{{Amazon-3M}} 
& {@1} 
&  {36.31$\pm$0.04} &  {36.31$\pm$0.04} 
&  {12.06$\pm$0.02} &  {12.06$\pm$0.02} 
&  {40.57} &  {40.57} 
&  {12.87} &  {12.87} 
\\ 
& {@3} 
&  {34.15$\pm$0.01} &  {34.99$\pm$0.01} 
&  {14.22$\pm$0.01} &  {13.67$\pm$0.01} 
&  {37.95} &  {38.96} 
&  {15.24} &  {14.63} 
\\ 
& {@5} 
&  {32.58$\pm$0.01} &  {34.12$\pm$0.01} 
&  {15.87$\pm$0.00} &  {14.83$\pm$0.01} 
&  {36.10} &  {37.91} 
&  {17.04} &  {15.90} 
\\ 
\hline 
\end{tabular} 
\end{center} 
\end{table*} 

In Table~\ref{tab:ErrorBounds}, we present the error bounds of the 
pairwise distances for different datasets by using 
$\epsilon = \sqrt{log(n)/d}$ (for simplicity, we omit the $O(\cdot)$ 
notation), where $n$ is the number of data 
points (training samples) and $d$ is the dimensionality of the 
input feature space. We consider the 
dimensionality of the (output) projection space as 
$r=200$, which is what 
we use to evaluate and compare the performance of OGEEC. 
Here, we observe that the error bounds 
($1-\epsilon$ and $1+\epsilon$) are comparable for all the 
datasets, and quite reasonable even when 
the dimensionality of the projection space is just $200$. 
To further investigate this, we vary the dimensionality of the 
projection space in $r=\{50,100,\ldots,400\}$ for the 
Delicious-200K dataset in Table~\ref{tab:VaryErrorBounds}. 
We observe that as we increase $r$, the value of $\epsilon$ 
reduces and becomes half as we move from $r=50$ to $r=200$ 
(from $0.3254$ to $0.1627$), however the rate of reduction 
slows down after that. 
As we will see later in our experiments, we observe similar 
trends in accuracies (Figure~\ref{fig:VaryEmbeddingDimension}) 
where they increase steeply in the 
beginning on increasing $r$ and then start levelling-out. 
Because of this, we 
fix the dimensionality of the output/projection space as 
$r=200$ in all the reported results, which also provides a 
reasonable prediction time for all the datasets.

\subsubsection{Evaluation Metrics} 
\label{sec:EvaluationMetrics} 

Following existing methods
methods~\cite{sleec,fastxml,pfastrexml,dismec,pdsparse,leml,parabel,proxml} 
and the evaluation metrics 
published on the XML repository~\cite{xmlrepo}, we use 
four metrics in our evaluations: Precision at $K$ (P@$K$), 
nDCG at $K$ (N@$K$), propensity-scored Precision at $K$ 
(PSP@$K$) and propensity-scored nDCG at $K$ 
(PSN@$K$), for $K \in \{1,3,5\}$. 
P@$K$ and PSP@$K$ metrics count the percentage of 
correctly predicted labels among the top $K$ labels, without 
considering the rank of correct labels among the 
predictions. N@$K$ and PSN@$K$ are ranking based 
measures and also take into account the position of correct 
labels among the top $K$ labels, with correct labels 
towards the top being considered better than those 
predicted towards the bottom of the top $K$ predictions. 
PSP@$K$ and PSN@$K$ were proposed in~\cite{pfastrexml}, 
which 
balance the correct prediction of rare and frequent labels 
by fitting a sigmoid function on the frequency distribution 
of the labels in the training set, and then use it to 
scale the  prediction scores of labels.

\subsubsection{Hyperparameters} 
\label{sec:HyperParameters} 

In all the main results and comparisons of OGEEC, we 
keep the dimensionality of the latent feature embedding 
space as 
$r=200$, the number of nearest neighbours in kNN as $k=5$, 
and an ensemble of five learners. 
These are chosen 
from a held-out validation set of the Delicious-200K 
dataset, and the same values 
are used for the other two data sets.

\subsection{Performance of OGEEC} 

\subsubsection{Results} 

Table~\ref{tab:ResultsNoProp} 
shows the performance of OGEEC in terms of 
all the four evaluation metrics using one learner 
and an ensemble of five learners. 
From these results, we can make the following observations: 
(1) As discussed earlier in 
Section~\ref{sec:Introduction}, even though the global 
feature embedding used in OGEEC is computed without 
using the 
training data, it achieves reasonable results on all the datasets, 
thus validating the promise of the JL-Lemma on the 
challenging XML task, and large-scale learning problems in 
general. 
(2) The variation in the performance of individual learners 
is statistically significant ($p$-value $<0.001$), with 
very small standard deviation. 
On using an ensemble of multiple learners, there is always 
a boost in the accuracy (by up to $\sim 4\%$ absolute in some cases). 
(3) The accuracy trends are similar across all the metrics and 
datasets, which demonstrate a consistent behaviour by OGEEC.

\begin{figure*}[h!] 
\begin{center}
\begin{tabular}{|c|c|}
\toprule 
  \includegraphics[width=0.45\textwidth]{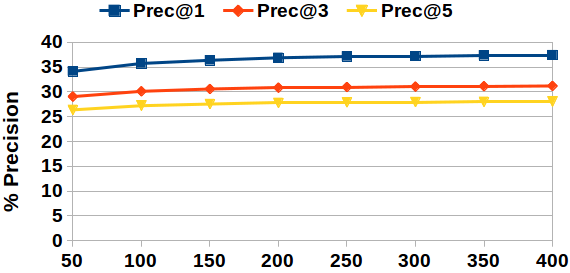} 
& \includegraphics[width=0.45\textwidth]{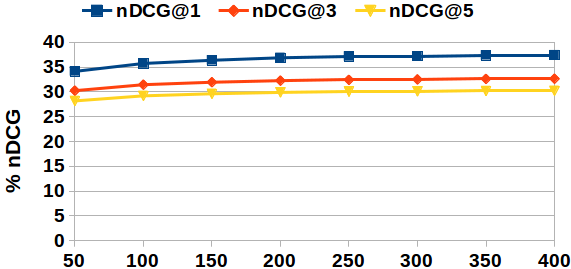} 
\\ 
\midrule 
  \includegraphics[width=0.45\textwidth]{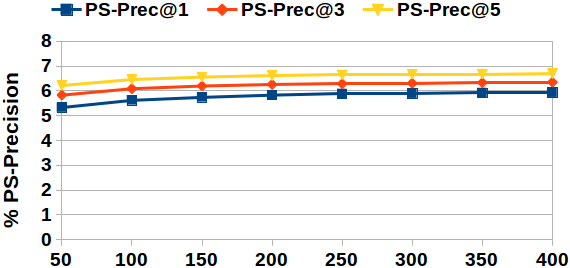} 
& \includegraphics[width=0.45\textwidth]{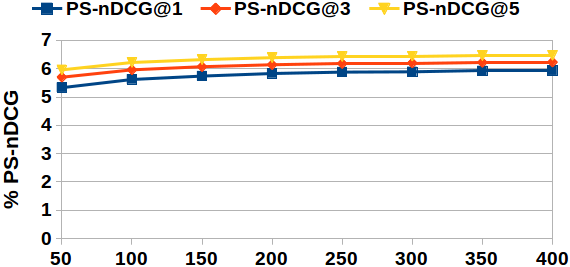} 
\\ 
\bottomrule 
\end{tabular}
\end{center}
\caption{Performance of OGEEC 
        on the Delicious-200K dataset 
        by varying the dimensionality of the 
        embedding space in $\{50,100,150,200, 250, 300, 350, 400\}$ 
        using one learner. 
}
\label{fig:VaryEmbeddingDimension} 
\end{figure*}

\begin{figure*}[h] 
\begin{center}
\begin{tabular}{|c|c|}
\toprule 
\includegraphics[width=0.45\textwidth]{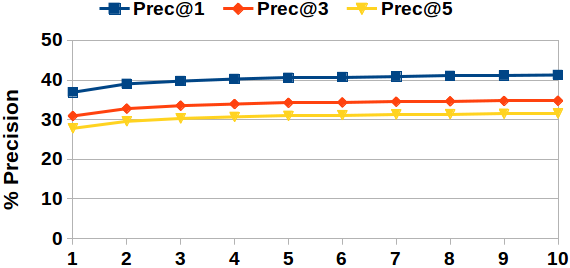} 
& \includegraphics[width=0.45\textwidth]{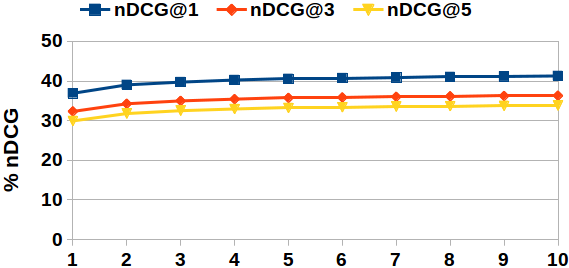} 
\\ 
\midrule 
\includegraphics[width=0.45\textwidth]{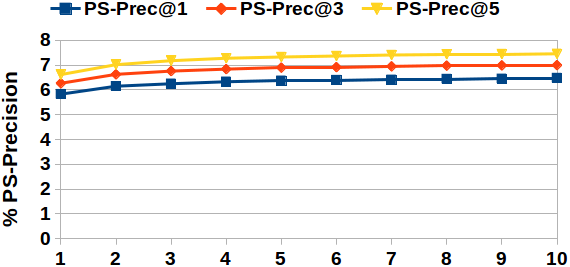} 
& \includegraphics[width=0.45\textwidth]{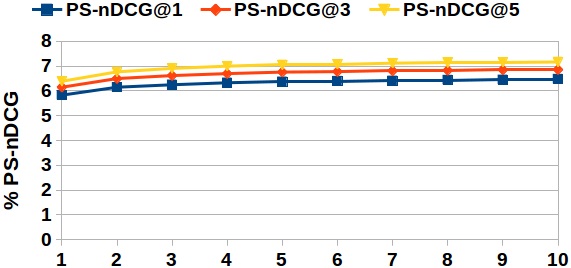} 
\\ 
\bottomrule 
\end{tabular}
\end{center}
\caption{Performance of OGEEC on the 
        Delicious-200K dataset 
        by varying the number of learners 
        in \{1, 2,\ldots,10\} 
        with $r=200$. 
}
\label{fig:VaryNoOfEnsembles}
\end{figure*}

\subsubsection{Ablation Studies} 

In Figure~\ref{fig:VaryEmbeddingDimension}, we study the 
influence of the dimensionality of the latent embedding 
space on the Delicious-200K 
dataset. To do so, we vary the embedding dimension 
in $\{50,100,\ldots,400\}$ and evaluate the performance of 
OGEEC using one learner. Here, we observe that 
the results using all the metrics 
consistently increase as we increase the 
embedding dimension, which is expected since the loss in 
information reduces with an increase in the number of dimensions. 
We also see that the improvements are steeper in the 
beginning and then gradually saturate. 
In Figure~\ref{fig:VaryNoOfEnsembles}, we study the 
influence of the number of learners in an ensemble 
by varying them in $\{1,2,\ldots,10\}$ 
on the Delicious-200K dataset. 
Here also, we observe that the accuracy increases 
sharply in the beginning and then starts 
saturating after training very few learners. 
In both these experiments, it is important to note that 
the computation cost increases with an increase in 
the dimensionality of the latent embedding space as well 
number of learners in an ensemble. 
Hence, based on the results in these experiments, we 
set the embedding dimension as $200$ and number of 
ensemble as $5$ in all our experiments 
to manage the trade-off between accuracies and 
computation cost.

\begin{table}
\caption{Computation time of OGEEC using 
        one learner. 
        Both training and prediction times are using 
        a single CPU core with $15.6$ GB RAM. 
} 
\label{tab:ComputationalCost}  
\begin{center} 
\begin{tabular}{ccc} 
\hline 
\multirow{2}{*}{ {Dataset}} 
& {\multirow{2}{*}{ {Training time (sec.)}}} 
& { {Prediction time}} 
\\ 
&& { {per sample (sec.)}} 
\\ 
\hline 
 {Delicious-200K} 
&  {2.46} 
&  {0.48} 
\\ 
 {Amazon-670K} 
&  {0.54} 
&  {0.48} 
\\ 
 {Amazon-3M} 
&  {1.14} 
&  {2.63} 
\\ 
\hline 
\end{tabular} 
\end{center} 
\end{table}

\begin{table}[t] 
\caption{Model details of OGEEC using one learner. 
        \# P: No. of parameters (= Feature embedding 
        matrix size); \# LP: No. of learned parameters; 
        \# HP: No. of hyper-parameters 
} 
\label{tab:ModelDetails} 
\begin{center} 
\begin{tabular}{cccc} 
\hline 
{ {Dataset}} 
&  {\# P} &  {\# LP} 
&  {\# HP} 
\\ 
\hline 
 {Delicious-200K} 
&  {782585 $\times$ 200} &  {0} 
&  {2} 
\\ 
 {Amazon-670K} 
&  {135909 $\times$ 200} &  {0} 
&  {2} 
\\ 
 {Amazon-3M} 
&  {337067 $\times$ 200} &  {0} 
&  {2} 
\\ 
\hline 
\end{tabular} 
\end{center} 
\end{table}

\begin{table}[htp] 
\caption{Comparison between LSH and OGEEC 
        in terms of prediction accuracies 
        and experimental requirements 
        on the Delicious-200K dataset 
} 
\label{tab:CompareLSH} 
\begin{center} 
\begin{tabular}{ccc} 
\hline 
\small{Metric} & \small{LSH} & \small{OGEEC (Ours)}
\\ 
\hline 
\small{P@1} & 38.90 & 40.54 
\\ 
\small{P@3} & 33.02 & 34.25 
\\ 
\small{P@5} & 30.07 & 30.97 
\\ 
\small{N@1} & 38.90 & 40.54 
\\ 
\small{N@3} & 34.40 & 35.74 
\\ 
\small{N@5} & 32.12 & 33.22 
\\ 
\small{PSP@1} & 6.04 & 6.37 
\\ 
\small{PSP@3} & 6.60 & 6.91 
\\ 
\small{PSP@5} & 7.07 & 7.33 
\\ 
\small{PSN@1} & 6.04 & 6.37 
\\ 
\small{PSN@3} & 6.45 & 6.76 
\\ 
\small{PSN@5} & 6.77 & 7.05 
\\ 
\small{Model Size (GB)} & 1.50 & 2.64 
\\ 
\small{Training Time (sec)} & 2.46 & 37.49 
\\ 
\small{Prediction Time (sec)} & 0.48 & 0.13 
\\ 
\hline 
\end{tabular} 
\end{center} 
\end{table}

\subsubsection{Computation time and Model details} 

We show the training and prediction time of OGEEC in 
Table~\ref{tab:ComputationalCost}, and the model details 
in terms of the number of parameters (which is also the 
size of the feature embedding matrix $\mathbf{F}$), the 
number of learned parameters, and the number of 
hyper-parameters in Table~\ref{tab:ModelDetails}. 
Here, the training and prediction time 
denote the time taken for processing the step(s) under 
``Method'' in the corresponding algorithm 
({\it c.f.} Algorithm-1 and Algorithm-2). 
The column ``\# Learned Parameters'' denotes the 
number of parameters in the embedding matrix 
that are learned using the training data, which is 
essentially zero in OGEEC. The significance 
of this is to specifically highlight that OGEEC requires 
only the dimensionality of the input feature space and the 
latent embedding space to obtain the projection matrix, 
without the need of training data. This is a unique 
property of OGEEC that makes it the first such technique 
in the XML literature as per our knowledge. 
From Table~\ref{tab:ComputationalCost}, 
we can notice that the prediction in OGEEC 
is also reasonably fast given the fact that 
we propagate labels after doing 
an exhaustive matching with all the 
training samples on just one CPU core (e.g., with 
$\sim 1.7$ million samples in the Amazon-3M dataset). 
In practice, the prediction time 
depends primarily on two factors: 
exact search of the nearest neighbours from the entire 
training set (which increases as the number of training samples 
increases), and propagation of labels from the identified 
nearest 
neighbours (which increases as the vocabulary size increases). 
At the time of deployment, we can reduce the prediction 
time significantly by first retrieving the nearest 
neighbours on subsets of training data in parallel, and then 
doing a second filtering in one pass 
after merging those small subsets. In a preliminary 
experiment, we observed that the prediction time of 
OGEEC on the Amazon-3M dataset can be brought down from 
$\sim2.6$ seconds per sample 
to an order of a few milliseconds without any compromise 
on accuracy by using 
a parallel version of the kNN algorithm on 24 CPU cores. 


\subsubsection{Comparison with Locality Sensitive Hashing} 
\label{sec:CompareWithLsh} 

Locality Sensitive Hashing (LSH) is a popular algorithm 
for fast and approximate search of nearest neighbours, and 
has been used in a variety of applications. 
As we perform an exhaustive kNN search in the proposed 
OGEEC approach, we compare it with LSH to analyze the 
advantages as well as limitations of adopting a simple 
technique. For fair comparisons, we learn the hash 
tables in LSH using the features in the latent embedding 
space of OGEEC. 
In Table-\ref{tab:CompareLSH}, first 
we compare the accuracies of OGEEC and LSH. 
Here, we observe that OGEEC usually achieves 
provide $\sim 3-15\%$ of relative improvements 
compared to LSH. 
Next, we can observe that while the model-size of 
both OGEEC and LSH are comparable, the training 
time of OGEEC is much less than that of LSH. 
Finally, we found that the prediction 
time of OGEEC and LSH were of the same order. 
From these comparisons, we can conclude that while 
the time requirements of both OGEEC and LSH are comparable 
for practical applications, 
OGEEC provides a clear empirical advantage 
over LSH.


\begin{table*}[htp] 
\caption{Comparison with non-deep learning based methods 
} 
\label{tab:CompareNonDeep} 
\begin{center} 
\begin{tabular}{cccccc} 
\cellcolor{MyColor}{Ours} 
& \cellcolor{MyColorLess}{Less than ours} 
& \cellcolor{MyColorMore1}{Better than ours by up to 3\%} 
& \cellcolor{MyColorMore2}{Better than ours by 3--6\%} 
& \cellcolor{MyColorMore3}{Better than ours by 6--9\%} 
& \cellcolor{MyColorMore4}{Better than ours by >9\%} 
\end{tabular} 
\begin{tabular}{ccccccccccccc} 
& & & & & & & & & & & & \\ 
\multicolumn{13}{c}{{\bf Delicious-200K}} 
\\ 
\hline 
\small{Method} 
& \small{P@1} & \small{P@3} & \small{P@5} 
& \small{N@1} & \small{N@3} & \small{N@5} 
& \small{PSP@1} & \small{PSP@3} & \small{PSP@5} 
&  \small{PSN@1} &  \small{PSN@3} &  \small{PSN@5} 
\\ 
\hline 
\small{OGEEC (ours)} 
& \cellcolor{MyColor}{40.54} & \cellcolor{MyColor}{34.25} & \cellcolor{MyColor}{30.97} 
& \cellcolor{MyColor}{40.54} & \cellcolor{MyColor}{35.74} & \cellcolor{MyColor}{33.22} 
& \cellcolor{MyColor}{6.37} & \cellcolor{MyColor}{6.91} & \cellcolor{MyColor}{7.33} 
& \cellcolor{MyColor}{6.37} & \cellcolor{MyColor}{6.76} & \cellcolor{MyColor}{7.05} 
\\ 
\small{LPSR-NB} 
& \cellcolor{MyColorLess}{18.59} & \cellcolor{MyColorLess}{15.43} & \cellcolor{MyColorLess}{14.07} 
& \cellcolor{MyColorLess}{18.59} & \cellcolor{MyColorLess}{16.17} & \cellcolor{MyColorLess}{15.13} 
& \cellcolor{MyColorLess}{3.24} & \cellcolor{MyColorLess}{3.42} & \cellcolor{MyColorLess}{3.64} 
& \cellcolor{MyColorLess}{3.24} & \cellcolor{MyColorLess}{3.37} & \cellcolor{MyColorLess}{3.52} 
\\ 
\small{PD-Sparse*} 
& \cellcolor{MyColorLess}{34.37} & \cellcolor{MyColorLess}{29.48} & \cellcolor{MyColorLess}{27.04} 
& \cellcolor{MyColorLess}{34.37} & \cellcolor{MyColorLess}{30.60} & \cellcolor{MyColorLess}{28.65} 
& \cellcolor{MyColorLess}{5.29} & \cellcolor{MyColorLess}{5.80} & \cellcolor{MyColorLess}{6.24} 
& \cellcolor{MyColorLess}{5.29} & \cellcolor{MyColorLess}{5.66} & \cellcolor{MyColorLess}{5.96} 
\\ 
\small{LEML*} 
& \cellcolor{MyColorMore1}{40.73} & \cellcolor{MyColorMore2}{37.71} & \cellcolor{MyColorMore2}{35.84} 
& \cellcolor{MyColorMore1}{40.73} & \cellcolor{MyColorMore1}{38.44} & \cellcolor{MyColorMore2}{37.01} 
& \cellcolor{MyColorLess}{6.06} & \cellcolor{MyColorMore1}{7.24} & \cellcolor{MyColorMore1}{8.10} 
& \cellcolor{MyColorLess}{6.06} & \cellcolor{MyColorMore1}{6.93} & \cellcolor{MyColorMore1}{7.52} 
\\ 
\small{PfastreXML*} 
& \cellcolor{MyColorMore1}{41.72} & \cellcolor{MyColorMore2}{37.83} & \cellcolor{MyColorMore2}{35.58} 
& \cellcolor{MyColorMore1}{41.72} & \cellcolor{MyColorMore2}{38.76} & \cellcolor{MyColorMore2}{37.08} 
& \cellcolor{MyColorLess}{3.15} & \cellcolor{MyColorLess}{3.87} & \cellcolor{MyColorLess}{4.43} 
& \cellcolor{MyColorLess}{3.15} & \cellcolor{MyColorLess}{3.68} & \cellcolor{MyColorLess}{4.06} 
\\ 
\small{FastXML*} 
& \cellcolor{MyColorMore1}{43.07} & \cellcolor{MyColorMore2}{38.66} & \cellcolor{MyColorMore2}{36.19} 
& \cellcolor{MyColorMore1}{43.07} & \cellcolor{MyColorMore2}{39.70} & \cellcolor{MyColorMore2}{37.83} 
& \cellcolor{MyColorMore1}{6.48} & \cellcolor{MyColorMore1}{7.52} & \cellcolor{MyColorMore1}{8.31} 
& \cellcolor{MyColorMore1}{6.51} & \cellcolor{MyColorMore1}{7.26} & \cellcolor{MyColorMore1}{7.79} 
\\ 
\small{DiSMEC*} 
& \cellcolor{MyColorMore2}{45.50} & \cellcolor{MyColorMore2}{38.70} & \cellcolor{MyColorMore2}{35.50} 
& \cellcolor{MyColorMore2}{45.50} & \cellcolor{MyColorMore2}{40.90} & \cellcolor{MyColorMore2}{37.80} 
& \cellcolor{MyColorMore1}{6.50} & \cellcolor{MyColorMore1}{7.60} & \cellcolor{MyColorMore1}{8.40} 
& \cellcolor{MyColorMore1}{6.50} & \cellcolor{MyColorMore1}{7.50} & \cellcolor{MyColorMore1}{7.90} 
\\ 
\small{Bonsai*} 
& \cellcolor{MyColorMore3}{46.69} & \cellcolor{MyColorMore2}{39.88} & \cellcolor{MyColorMore2}{36.38} 
& \cellcolor{MyColorMore3}{46.69} & \cellcolor{MyColorMore2}{41.51} & \cellcolor{MyColorMore2}{38.84} 
& \cellcolor{MyColorMore1}{7.26} & \cellcolor{MyColorMore1}{7.97} & \cellcolor{MyColorMore1}{8.53} 
& \cellcolor{MyColorMore1}{7.26} & \cellcolor{MyColorMore1}{7.75} & \cellcolor{MyColorMore1}{8.10} 
\\ 
\small{AnnexML*} 
& \cellcolor{MyColorMore3}{46.79} & \cellcolor{MyColorMore3}{40.72} & \cellcolor{MyColorMore3}{37.67} 
& \cellcolor{MyColorMore3}{46.79} & \cellcolor{MyColorMore3}{42.17} & \cellcolor{MyColorMore3}{39.84} 
& \cellcolor{MyColorMore1}{7.18} & \cellcolor{MyColorMore1}{8.05} & \cellcolor{MyColorMore1}{8.74} 
& \cellcolor{MyColorMore1}{7.18} & \cellcolor{MyColorMore1}{7.78} & \cellcolor{MyColorMore1}{8.22} 
\\ 
\small{Parabel*} 
& \cellcolor{MyColorMore3}{46.86} & \cellcolor{MyColorMore3}{40.08} & \cellcolor{MyColorMore2}{36.70} 
& \cellcolor{MyColorMore3}{46.86} & \cellcolor{MyColorMore2}{41.69} & \cellcolor{MyColorMore2}{39.10} 
& \cellcolor{MyColorMore1}{7.22} & \cellcolor{MyColorMore1}{7.94} & \cellcolor{MyColorMore1}{8.54} 
& \cellcolor{MyColorMore1}{7.22} & \cellcolor{MyColorMore1}{7.71} & \cellcolor{MyColorMore1}{8.09} 
\\ 
\small{SLEEC*} 
& \cellcolor{MyColorMore3}{47.85} & \cellcolor{MyColorMore3}{42.21} & \cellcolor{MyColorMore3}{39.43} 
& \cellcolor{MyColorMore3}{47.85} & \cellcolor{MyColorMore3}{43.52} & \cellcolor{MyColorMore3}{41.37} 
& \cellcolor{MyColorMore1}{7.17} & \cellcolor{MyColorMore1}{8.16} & \cellcolor{MyColorMore1}{8.96} 
& \cellcolor{MyColorMore1}{7.17} & \cellcolor{MyColorMore1}{7.89} & \cellcolor{MyColorMore1}{8.44} 
\\ 
\hline 
& & & & & & & & & & & & \\ 
\multicolumn{13}{c}{{\bf Amazon-670K}} 
\\ 
\hline 
\small{Method} 
& \small{P@1} & \small{P@3} & \small{P@5} 
& \small{N@1} & \small{N@3} & \small{N@5} 
& \small{PSP@1} & \small{PSP@3} & \small{PSP@5} 
&  \small{PSN@1} &  \small{PSN@3} &  \small{PSN@5} 
\\ 
\hline 
\small{OGEEC (ours)} 
& \cellcolor{MyColor}{37.45} & \cellcolor{MyColor}{33.67} & \cellcolor{MyColor}{31.12} 
& \cellcolor{MyColor}{37.45} & \cellcolor{MyColor}{35.64} & \cellcolor{MyColor}{34.68} 
& \cellcolor{MyColor}{23.05} & \cellcolor{MyColor}{26.19} & \cellcolor{MyColor}{29.42} 
& \cellcolor{MyColor}{23.05} & \cellcolor{MyColor}{25.37} & \cellcolor{MyColor}{27.54} 
\\ 
\small{LEML*} 
& \cellcolor{MyColorLess}{8.13} & \cellcolor{MyColorLess}{6.83} & \cellcolor{MyColorLess}{6.03} 
& \cellcolor{MyColorLess}{8.13} & \cellcolor{MyColorLess}{7.30} & \cellcolor{MyColorLess}{6.85} 
& \cellcolor{MyColorLess}{2.07} & \cellcolor{MyColorLess}{2.26} & \cellcolor{MyColorLess}{2.47} 
& \cellcolor{MyColorLess}{2.07} & \cellcolor{MyColorLess}{2.21} & \cellcolor{MyColorLess}{2.35} 
\\ 
\small{LPSR-NB*} 
& \cellcolor{MyColorLess}{28.65} & \cellcolor{MyColorLess}{24.88} & \cellcolor{MyColorLess}{22.37} 
& \cellcolor{MyColorLess}{28.65} & \cellcolor{MyColorLess}{26.40} & \cellcolor{MyColorLess}{25.03} 
& \cellcolor{MyColorLess}{16.68} & \cellcolor{MyColorLess}{18.07} & \cellcolor{MyColorLess}{19.43} 
& \cellcolor{MyColorLess}{16.68} & \cellcolor{MyColorLess}{17.70} & \cellcolor{MyColorLess}{18.63} 
\\ 
\small{SLICE+FastText*} 
& \cellcolor{MyColorLess}{33.15} & \cellcolor{MyColorLess}{29.76} & \cellcolor{MyColorLess}{26.93} 
& \cellcolor{MyColorLess}{33.15} & \cellcolor{MyColorLess}{31.51} & \cellcolor{MyColorLess}{30.27} 
& \cellcolor{MyColorLess}{20.20} & \cellcolor{MyColorLess}{22.69} & \cellcolor{MyColorLess}{24.70} 
& \cellcolor{MyColorLess}{20.20} & \cellcolor{MyColorLess}{21.71} & \cellcolor{MyColorLess}{22.72} 
\\ 
\small{SLEEC*} 
& \cellcolor{MyColorLess}{35.05} & \cellcolor{MyColorLess}{31.25} & \cellcolor{MyColorLess}{28.56} 
& \cellcolor{MyColorLess}{34.77} & \cellcolor{MyColorLess}{32.74} & \cellcolor{MyColorLess}{31.53} 
& \cellcolor{MyColorLess}{20.62} & \cellcolor{MyColorLess}{23.32} & \cellcolor{MyColorLess}{25.98} 
& \cellcolor{MyColorLess}{20.62} & \cellcolor{MyColorLess}{22.63} & \cellcolor{MyColorLess}{24.43} 
\\ 
\small{FastXML*} 
& \cellcolor{MyColorLess}{36.99} & \cellcolor{MyColorLess}{33.28} & \cellcolor{MyColorLess}{30.53} 
& \cellcolor{MyColorLess}{36.99} & \cellcolor{MyColorLess}{35.11} & \cellcolor{MyColorLess}{33.86} 
& \cellcolor{MyColorLess}{19.37} & \cellcolor{MyColorLess}{23.26} & \cellcolor{MyColorLess}{26.85} 
& \cellcolor{MyColorLess}{19.37} & \cellcolor{MyColorLess}{22.25} & \cellcolor{MyColorLess}{24.69} 
\\ 
\small{PfastreXML*} 
& \cellcolor{MyColorMore1}{39.46} & \cellcolor{MyColorMore1}{35.81} & \cellcolor{MyColorMore1}{33.05} 
& \cellcolor{MyColorMore1}{39.46} & \cellcolor{MyColorMore1}{37.78} & \cellcolor{MyColorMore1}{36.69} 
& \cellcolor{MyColorMore3}{29.30} & \cellcolor{MyColorMore2}{30.80} & \cellcolor{MyColorMore2}{32.43} 
& \cellcolor{MyColorMore3}{29.30} & \cellcolor{MyColorMore2}{30.40} & \cellcolor{MyColorMore2}{31.49} 
\\ 
\small{AnnexML*} 
& \cellcolor{MyColorMore2}{42.39} &  \cellcolor{MyColorMore2}{36.89} &  \cellcolor{MyColorMore1}{32.98} 
& \cellcolor{MyColorMore2}{42.39} &  \cellcolor{MyColorMore2}{39.07} &  \cellcolor{MyColorMore1}{37.04} 
& \cellcolor{MyColorLess}{21.56} & \cellcolor{MyColorLess}{24.78} & \cellcolor{MyColorLess}{27.66} 
& \cellcolor{MyColorLess}{21.56} & \cellcolor{MyColorLess}{23.38} & \cellcolor{MyColorLess}{24.76} 
\\ 
\small{ProXML*} 
& \cellcolor{MyColorMore3}{43.50} & \cellcolor{MyColorMore2}{38.70} & \cellcolor{MyColorMore2}{35.30} 
& \cellcolor{MyColorMore3}{43.50} & \cellcolor{MyColorMore2}{41.10} & \cellcolor{MyColorMore2}{39.70} 
& \cellcolor{MyColorMore3}{30.80} & \cellcolor{MyColorMore3}{32.80} & \cellcolor{MyColorMore2}{35.10} 
& \cellcolor{MyColorMore3}{30.80} & \cellcolor{MyColorMore3}{31.70} & \cellcolor{MyColorMore2}{32.70} 
\\ 
\small{DiSMEC*} 
& \cellcolor{MyColorMore3}{44.70} & \cellcolor{MyColorMore3}{39.70} & \cellcolor{MyColorMore2}{36.10} 
& \cellcolor{MyColorMore3}{44.70} & \cellcolor{MyColorMore3}{42.10} & \cellcolor{MyColorMore2}{40.50} 
& \cellcolor{MyColorMore2}{27.80} & \cellcolor{MyColorMore2}{30.60} & \cellcolor{MyColorMore2}{34.20} 
& \cellcolor{MyColorMore2}{27.80} & \cellcolor{MyColorMore2}{28.80} & \cellcolor{MyColorMore2}{30.70} 
\\ 
\small{Parabel*} 
& \cellcolor{MyColorMore3}{44.89} & \cellcolor{MyColorMore3}{39.80} & \cellcolor{MyColorMore2}{36.00} 
& \cellcolor{MyColorMore3}{44.89} & \cellcolor{MyColorMore3}{42.14} & \cellcolor{MyColorMore2}{40.36} 
& \cellcolor{MyColorMore1}{25.43} & \cellcolor{MyColorMore2}{29.43} & \cellcolor{MyColorMore2}{32.85} 
& \cellcolor{MyColorMore1}{25.43} & \cellcolor{MyColorMore2}{28.38} & \cellcolor{MyColorMore2}{30.71} 
\\ 
\small{PPD-Sparse*} 
& \cellcolor{MyColorMore3}{45.32} & \cellcolor{MyColorMore3}{40.37} & \cellcolor{MyColorMore2}{36.92} 
& -- & -- & -- 
& \cellcolor{MyColorMore2}{26.64} & \cellcolor{MyColorMore2}{30.65} & \cellcolor{MyColorMore2}{34.65} 
& -- & -- & -- 
\\ 
\small{Bonsai*} 
& \cellcolor{MyColorMore3}{45.58} & \cellcolor{MyColorMore3}{40.39} & \cellcolor{MyColorMore2}{36.60} 
& \cellcolor{MyColorMore3}{45.58} & \cellcolor{MyColorMore3}{42.79} & \cellcolor{MyColorMore3}{41.05} 
& \cellcolor{MyColorMore2}{27.08} & \cellcolor{MyColorMore2}{30.79} & \cellcolor{MyColorMore2}{34.11} 
& -- & -- & -- 
\\ 
\hline 
& & & & & & & & & & & & \\ 
\multicolumn{13}{c}{{\bf Amazon-3M}} 
\\ 
\hline 
\small{Method} 
& \small{P@1} & \small{P@3} & \small{P@5} 
& \small{N@1} & \small{N@3} & \small{N@5} 
& \small{PSP@1} & \small{PSP@3} & \small{PSP@5} 
&  \small{PSN@1} &  \small{PSN@3} &  \small{PSN@5} 
\\ 
\hline 
\small{OGEEC (ours)} 
& \cellcolor{MyColor}{40.57} & \cellcolor{MyColor}{37.95} & \cellcolor{MyColor}{36.10} 
& \cellcolor{MyColor}{40.57} & \cellcolor{MyColor}{38.96} & \cellcolor{MyColor}{37.91} 
& \cellcolor{MyColor}{12.87} & \cellcolor{MyColor}{15.24} & \cellcolor{MyColor}{17.04} 
& \cellcolor{MyColor}{12.87} & \cellcolor{MyColor}{14.63} & \cellcolor{MyColor}{15.90} 
\\ 
\small{PfastreXML*} 
& \cellcolor{MyColorMore2}{43.83} & \cellcolor{MyColorMore2}{41.81} & \cellcolor{MyColorMore2}{40.09} 
& \cellcolor{MyColorMore2}{43.83} & \cellcolor{MyColorMore2}{42.68} & \cellcolor{MyColorMore2}{41.75} 
& \cellcolor{MyColorMore3}{21.38} & \cellcolor{MyColorMore3}{23.22} & \cellcolor{MyColorMore3}{24.52} 
& \cellcolor{MyColorMore3}{21.38} & \cellcolor{MyColorMore3}{22.75} & \cellcolor{MyColorMore3}{23.68} 
\\ 
\small{FastXML*} 
& \cellcolor{MyColorMore2}{44.24} & \cellcolor{MyColorMore1}{40.83} & \cellcolor{MyColorMore1}{38.59} 
& \cellcolor{MyColorMore2}{44.24} & \cellcolor{MyColorMore1}{41.92} & \cellcolor{MyColorMore1}{40.47} 
& \cellcolor{MyColorLess}{9.77} & \cellcolor{MyColorLess}{11.69} &  \cellcolor{MyColorLess}{13.25} 
& \cellcolor{MyColorLess}{9.77} & \cellcolor{MyColorLess}{11.20} &  \cellcolor{MyColorLess}{12.29} 
\\ 
\small{DiSMEC} 
& \cellcolor{MyColorMore3}{47.34} & \cellcolor{MyColorMore3}{44.96} & \cellcolor{MyColorMore3}{42.80} 
& \cellcolor{MyColorMore3}{47.36} & -- &  -- 
& -- &  -- &  -- 
& -- &  -- &  -- 
\\ 
\small{Parabel*} 
& \cellcolor{MyColorMore3}{47.48} & \cellcolor{MyColorMore3}{44.65} & \cellcolor{MyColorMore3}{42.53} 
& \cellcolor{MyColorMore3}{47.48} & \cellcolor{MyColorMore3}{45.73} & \cellcolor{MyColorMore3}{44.53} 
& \cellcolor{MyColorLess}{12.82} & \cellcolor{MyColorMore1}{15.61} & \cellcolor{MyColorMore1}{17.73} 
& \cellcolor{MyColorLess}{12.82} &  \cellcolor{MyColorMore1}{14.89} &  \cellcolor{MyColorMore1}{16.38} 
\\ 
\small{Bonsai*} 
& \cellcolor{MyColorMore3}{48.45} & \cellcolor{MyColorMore3}{45.65} & \cellcolor{MyColorMore3}{43.49} 
& \cellcolor{MyColorMore3}{48.45} & \cellcolor{MyColorMore3}{46.78} & \cellcolor{MyColorMore3}{45.59} 
& \cellcolor{MyColorMore1}{13.79} &  \cellcolor{MyColorMore1}{16.71} &  \cellcolor{MyColorMore1}{18.87} 
& -- &  -- &  -- 
\\ 
\small{AnnexML*} 
& \cellcolor{MyColorMore3}{49.30} & \cellcolor{MyColorMore3}{45.55} & \cellcolor{MyColorMore3}{43.11} 
& \cellcolor{MyColorMore3}{49.30} & \cellcolor{MyColorMore3}{46.79} & \cellcolor{MyColorMore3}{45.27} 
& \cellcolor{MyColorLess}{11.69}  & \cellcolor{MyColorLess}{14.07} &  \cellcolor{MyColorLess}{15.98} 
& -- &  -- &  --
\\ 
\hline 
\end{tabular} 
\end{center} 
\end{table*}

\subsection{Comparison with the state-of-the-art} 
\label{sec:ComparisonWithSota} 

Now we demonstrate the efficiency and effectiveness of 
the proposed OGEEC approach by comparing it with several 
non-deep learning based as well deep learning based 
state-of-the-art XML methods in terms of accuracies, 
model-size and training time. 
Among the non-deep learning based methods, we compare 
with LPSR-NB~\cite{lpsr}, LEML~\cite{leml}, 
SLEEC~\cite{sleec}, AnnexML~\cite{annexml}, 
PD-Sparse~\cite{pdsparse}, PPD-Sparse~\cite{ppdsparse}, 
DiSMEC~\cite{dismec}, ProXML~\cite{proxml}, 
FastXML~\cite{fastxml}, PfastreXML~\cite{pfastrexml}, 
Parabel~\cite{parabel}, 
Bonsai~\cite{bonsai} and SLICE~\cite{slice}. 
Among the deep learning based methods, we compare 
with XML-CNN~\cite{xmlcnn}, DeepXML~\cite{deepxml}, 
APLC-XLNet~\cite{aplcxlnet}, Astec~\cite{astec} 
X-Transformer~\cite{xtransformer}, 
and AttentionXML~\cite{attentionxml}. 
For benchmarking, we adopt the results published on 
the XML repository~\cite{xmlrepo}.\footnote{Retrieved 
on 13 October, 2021.}. 
For ease of comparisons, we highlight 
the results in different colors. 

We first compare with non-deep learning based methods 
in Table~\ref{tab:CompareNonDeep}. 
In general, we notice that despite its simplicity, 
OGEEC achieves quite 
competitive results, which are many a times even better, 
compared to other methods all of which are 
algorithmically much more complex than it. 
We can also make several other observations from 
these results: 
(1) The accuracy of OGEEC in terms of propensity-scored 
metrics is particularly impressive, where it outperforms 
many of the existing methods. This indicates that our 
initial hypothesis of posing XML as a retrieval task 
and propagating labels from globally identified 
nearest neighbours in a low-dimensional 
structure-preserving space 
is in fact quite effective in boosting 
the accuracy of low-frequency labels. It should be 
noted that while the existing benchmark 
nearest-neighbour 
based approaches such as LEML~\cite{leml}, 
SLEEC~\cite{sleec} and AnnexML~\cite{annexml} 
have advocated the use of multiple local feature embeddings, 
ours is the first attempt that demonstrates the 
effectiveness of using a global structure-preserving 
feature embedding in the XML task as per our knowledge. 
(2) Similarly, using non-propensity scored measures, 
the accuracy of OGEEC is quite impressive. 
For most of the cases where OGEEC is outperformed 
by other methods, the difference in accuracy is below 
$6\%$ despite elaborate modelling and training efforts. 
We acknowledge that a performance gap by a few 
percentage might be critical in certain applications, however 
comparisons with a simple approach like 
OGEEC would help in better understanding and appreciating 
the benefits of the sophisticated
modelling and training procedures used by the existing
methods. 
(3) Among the existing embedding based methods, only 
AnnexML could scale to the largest Amazon-3M dataset 
Similarly, among the linear one-vs.-all classifier based 
approaches such as DiSMEC, ProXML, PD-Sparse and PPD-Sparse, only 
DiSMEC could be scaled on this dataset by using 
multiple CPU core (if we had used one CPU core, the 
training time of DiSMEC would have been $\approx 4995$ hours). 
Only the tree-based approaches such as FastXML, PfastreXML 
and Parabel that partition the 
label-space in a hierarchical manner could manage to scale 
to this dataset (e.g., the training time of PfastreXML is 
$\approx15$ hours using $16$ CPU cores). 
On the other hand, OGEEC, being an embedding-based approach, 
easily scales to this dataset and 
is able to generate an ensemble of five learners in just few 
seconds ({\it c.f.} Table~\ref{tab:ComputationalCost}) 
using just one CPU core and $15.6$ GB RAM. 
(4) On the Amazon-3M dataset, the performance 
of OGEEC is quite reasonable. In terms of P@K and 
N@K, it is quite close to FastXML and PfastreXML, and 
competitive to other methods. 
In terms of PSP@K and PSN@K, the accuracy of OGEEC is better than 
FastXML and AnnexML, comparable to Parabel and Bonsai, and 
competitive to PfastreXML. 
Here, it is worth noting that AnnexML is the current state-of-the-art 
non-deep learning based embedding approach for XML, and is the 
closest competitor to OGEEC.

\begin{table*}[htp] 
\caption{Comparison with deep learning based methods 
} 
\label{tab:CompareDeep} 
\begin{center} 
\begin{tabular}{cccccc} 
\cellcolor{MyColor}{Ours} 
& \cellcolor{MyColorLess}{Less than ours} 
& \cellcolor{MyColorMore1}{Better than ours by up to 3\%} 
& \cellcolor{MyColorMore2}{Better than ours by 3--6\%} 
& \cellcolor{MyColorMore3}{Better than ours by 6--9\%} 
& \cellcolor{MyColorMore4}{Better than ours by >9\%} 
\end{tabular} 
\begin{tabular}{ccccccccccccc} 
& & & & & & & & & & & & \\ 
\multicolumn{13}{c}{{\bf Delicious-200K}} 
\\ 
\hline 
\small{Method} 
& \small{P@1} & \small{P@3} & \small{P@5} 
& \small{N@1} & \small{N@3} & \small{N@5} 
& \small{PSP@1} & \small{PSP@3} & \small{PSP@5} 
&  \small{PSN@1} &  \small{PSN@3} &  \small{PSN@5} 
\\ 
\hline 
\small{OGEEC (ours)} 
& \cellcolor{MyColor}{40.54} & \cellcolor{MyColor}{34.25} & \cellcolor{MyColor}{30.97} 
& \cellcolor{MyColor}{40.54} & \cellcolor{MyColor}{35.74} & \cellcolor{MyColor}{33.22} 
& \cellcolor{MyColor}{6.37} & \cellcolor{MyColor}{6.91} & \cellcolor{MyColor}{7.33} 
& \cellcolor{MyColor}{6.37} & \cellcolor{MyColor}{6.76} & \cellcolor{MyColor}{7.05} 
\\ 
\small{X-Transformer*} 
& \cellcolor{MyColorMore2}{45.59} & \cellcolor{MyColorMore2}{39.10} & \cellcolor{MyColorMore2}{35.92} 
& \cellcolor{MyColorMore2}{45.59} & \cellcolor{MyColorMore2}{40.62} & \cellcolor{MyColorMore2}{38.17} 
& \cellcolor{MyColorMore1}{6.96} & \cellcolor{MyColorMore1}{7.71} & \cellcolor{MyColorMore1}{8.33} 
& \cellcolor{MyColorMore1}{6.96} & \cellcolor{MyColorMore1}{7.47} & \cellcolor{MyColorMore1}{7.86} 
\\ 
\hline 
& & & & & & & & & & & & \\ 
\multicolumn{13}{c}{{\bf Amazon-670K}} 
\\ 
\hline 
\small{Method} 
& \small{P@1} & \small{P@3} & \small{P@5} 
& \small{N@1} & \small{N@3} & \small{N@5} 
& \small{PSP@1} & \small{PSP@3} & \small{PSP@5} 
&  \small{PSN@1} &  \small{PSN@3} &  \small{PSN@5} 
\\ 
\hline 
\small{OGEEC (ours)} 
& \cellcolor{MyColor}{37.45} & \cellcolor{MyColor}{33.67} & \cellcolor{MyColor}{31.12} 
& \cellcolor{MyColor}{37.45} & \cellcolor{MyColor}{35.64} & \cellcolor{MyColor}{34.68} 
& \cellcolor{MyColor}{23.05} & \cellcolor{MyColor}{26.19} & \cellcolor{MyColor}{29.42} 
& \cellcolor{MyColor}{23.05} & \cellcolor{MyColor}{25.37} & \cellcolor{MyColor}{27.54} 
\\ 
\small{XML-CNN$\blacklozenge$} 
& \cellcolor{MyColorLess}{35.39} & \cellcolor{MyColorLess}{31.93} & \cellcolor{MyColorLess}{29.32} 
& \cellcolor{MyColorLess}{35.39} & \cellcolor{MyColorLess}{33.74} & \cellcolor{MyColorLess}{32.64} 
& \cellcolor{MyColorMore2}{28.67} & \cellcolor{MyColorMore3}{33.27} & \cellcolor{MyColorMore3}{36.51} 
& {--} & {--} & {--} 
\\ 
\small{DeepXML$\blacklozenge$} 
& \cellcolor{MyColorMore1}{37.67} & \cellcolor{MyColorMore1}{33.72} & \cellcolor{MyColorLess}{29.86} 
& \cellcolor{MyColorMore1}{37.67} & \cellcolor{MyColorLess}{33.93} & \cellcolor{MyColorLess}{32.47} 
& -- & -- & -- & -- & -- & -- 
\\ 
\small{X-Transformer*} 
& \cellcolor{MyColorMore2}{42.50} & \cellcolor{MyColorMore2}{37.87} & \cellcolor{MyColorMore2}{34.41} 
& \cellcolor{MyColorMore2}{42.50} & \cellcolor{MyColorMore2}{40.01} & \cellcolor{MyColorMore2}{38.43} 
& \cellcolor{MyColorMore1}{24.82} & \cellcolor{MyColorMore1}{28.20} & \cellcolor{MyColorMore1}{31.24} 
& \cellcolor{MyColorMore1}{24.82} & \cellcolor{MyColorMore1}{26.82} & \cellcolor{MyColorMore1}{28.29} 
\\ 
\small{APLC-XLNet$\blacklozenge$} 
& \cellcolor{MyColorMore3}{43.46} & \cellcolor{MyColorMore2}{38.83} & \cellcolor{MyColorMore2}{35.32} 
& \cellcolor{MyColorMore3}{43.46} & \cellcolor{MyColorMore2}{41.01} & \cellcolor{MyColorMore2}{39.38} 
& \cellcolor{MyColorMore2}{26.12} & \cellcolor{MyColorMore2}{29.66} & \cellcolor{MyColorMore2}{32.78} 
& \cellcolor{MyColorMore2}{26.12} & \cellcolor{MyColorMore1}{28.20} & \cellcolor{MyColorMore1}{29.68} 
\\ 
\small{AttentionXML$\blacklozenge$} 
& \cellcolor{MyColorMore4}{47.58} & \cellcolor{MyColorMore3}{42.61} & \cellcolor{MyColorMore3}{38.92} 
& \cellcolor{MyColorMore4}{47.58} & \cellcolor{MyColorMore4}{45.07} & \cellcolor{MyColorMore3}{43.50} 
& \cellcolor{MyColorMore3}{30.29} & \cellcolor{MyColorMore3}{33.85} & \cellcolor{MyColorMore3}{37.13} 
& -- & -- & -- 
\\ 
\small{Astec$\dagger$} 
& \cellcolor{MyColorMore4}{47.77} & \cellcolor{MyColorMore4}{42.79} & \cellcolor{MyColorMore3}{39.10} 
& \cellcolor{MyColorMore4}{47.77} & \cellcolor{MyColorMore4}{45.28} & \cellcolor{MyColorMore4}{43.74} 
& \cellcolor{MyColorMore4}{32.13} & \cellcolor{MyColorMore3}{35.14} & \cellcolor{MyColorMore3}{37.82} 
& \cellcolor{MyColorMore4}{32.13} & \cellcolor{MyColorMore3}{33.80} & \cellcolor{MyColorMore3}{35.01} 
\\ 
\hline 
& & & & & & & & & & & & \\ 
\multicolumn{13}{c}{{\bf Amazon-3M}} 
\\ 
\hline 
\small{Method} 
& \small{P@1} & \small{P@3} & \small{P@5} 
& \small{N@1} & \small{N@3} & \small{N@5} 
& \small{PSP@1} & \small{PSP@3} & \small{PSP@5} 
&  \small{PSN@1} &  \small{PSN@3} &  \small{PSN@5} 
\\ 
\hline 
\small{OGEEC (ours)} 
& \cellcolor{MyColor}{40.57} & \cellcolor{MyColor}{37.95} & \cellcolor{MyColor}{36.10} 
& \cellcolor{MyColor}{40.57} & \cellcolor{MyColor}{38.96} & \cellcolor{MyColor}{37.91} 
& \cellcolor{MyColor}{12.87} & \cellcolor{MyColor}{15.24} & \cellcolor{MyColor}{17.04} 
& \cellcolor{MyColor}{12.87} & \cellcolor{MyColor}{14.63} & \cellcolor{MyColor}{15.90} 
\\ 
\small{AttentionXML$\dagger$} 
& \cellcolor{MyColorMore4}{50.86} & \cellcolor{MyColorMore4}{48.04} & \cellcolor{MyColorMore4}{45.83} 
& \cellcolor{MyColorMore4}{50.86} & \cellcolor{MyColorMore4}{49.16} & \cellcolor{MyColorMore4}{47.94} 
& \cellcolor{MyColorMore1}{15.52} & \cellcolor{MyColorMore2}{18.45} & \cellcolor{MyColorMore2}{20.60} 
& -- &  -- &  -- 
\\ 
\hline 
\end{tabular} 
\end{center} 
\end{table*}


In Table~\ref{tab:CompareDeep}, we compare OGEEC 
with deep learning based XML methods. 
Among these, XML-CNN~\cite{xmlcnn} and DeepXML~\cite{deepxml} 
can be considered as baseline deep methods that 
were among the initial methods to explore the 
applicability of deep learning in XML. 
Whereas recent methods such as AttentionXML~\cite{attentionxml}, 
X-Transformer~\cite{xtransformer} use heavily 
parametrized and compute-intensive attention-based deep 
models using a cluster of multiple GPUs 
(their training times are scaled accordingly). 
As we can see from the results, the proposed OGEEC approach 
achieves results that are either better than or comparable 
to both XML-CNN and DeepXML. 
Analogous to our previous observation while comparing 
OGEEC with non-deep learning based methods, here also for 
most of the cases where OGEEC is outperformed 
by other deep methods, the difference in accuracy is below 
$6\%$ despite intensive modelling and training. 
In case of the largest Amazon-3M dataset on which only 
AttentionXML has been shown to scale, the performance of OGEEC is 
within $\approx3\%$ in terms of propensity-score metrics, and 
quite appealing in terms of other metrics given its 
interesting properties.


We compare the training time and model-size 
of existing methods with OGEEC in Table~\ref{tab:CompareMemoryTime}. 
In comparison to non-deep learning based approaches, 
we can observe that both training time 
and model-size of OGEEC are significantly less 
compared to all other methods. 
Specifically on the Amazon-3M 
dataset on which only a few existing XML methods 
have been able to scale, OGEEC provides 
$\approx6572\times$ speed-up ratio compared to 
the second fastest method Parabel, and is 
$\approx3.14$ Million times faster than the slowest DiSMEC. 
In terms of model size, OGEEC is around 
$26\times$, $14.7\times$ and $16x\times$ 
lighter than Parabel, PfastreXML and DiSMEC respectively. 
In comparison to deep learning based approaches as well, 
we can make similar observations; e.g., 
the memory footprint of OGEEC is 10-20 times smaller compared to 
AttentionXML while being 1000-100000 times faster. 
Here, it is important to recall that all these methods 
except OGEEC are trained learners and need to store 
the learnt models, however the learners in 
OGEEC can be obtained {\it on-the-fly} without using the 
training data and thus one may not require to store them 
explicitly. 
The above comparisons indicate that for 
extremely high-dimensional and large-scale 
data that is widely used in industrial applications, the 
performance of the OGEEC approach, that relies on 
inherrently non-linear and structure-preserving embeddings, 
can be quite competitive to other methods that are 
highly resource-intensive and learn up to a few 
billions of parameters. 

\begin{table}[htp] 
\caption{Comparison with existing XML methods in terms of model size and 
training time (note that one CPU core is used for OGEEC) 
\{$\dagger$: 24 CPU cores; 
$\ddagger$: 24 CPU cores with 1 Nvidia P40 GPU; 
$\blacklozenge$: Results as reported in publication\} 
} 
\label{tab:CompareMemoryTime} 
\begin{center} 
\begin{tabular}{ccc} 
\multicolumn{3}{c}{{\bf Delicious-200K}} 
\\ 
\hline 
\small{Method} 
& \small{Model Size (GB)} & \small{Training time (hr)} 
\\ 
\hline 
\small{OGEEC (ours)} 
& 1.50 & 0.004 
\\ 
\small{PfastreXML$\dagger$} 
& 15.34 & 3.60 
\\ 
\small{Bonsai$\dagger$} 
& 3.91 & 64.42 
\\ 
\small{AnnexML$\dagger$} 
& 10.74 & 2.58 
\\ 
\small{Parabel$\dagger$} 
& 6.36 & 9.58 
\\ 
\small{X-Transformer$\ddagger$} 
& 2.70 & 31.22 
\\ 
\hline 
& & \\ 
\multicolumn{3}{c}{{\bf Amazon-670K}} 
\\ 
\hline 
\small{Method} 
& \small{Model Size (GB)} & \small{Training time (hr)} 
\\ 
\hline 
\small{OGEEC (ours)} 
& {1.01} & {0.001} 
\\ 
\small{SLICE+FastText$\dagger$} 
& 2.01 & 0.21 
\\ 
\small{SLEEC$\dagger$} 
& 7.08 & 11.33 
\\ 
\small{PfastreXML$\dagger$} 
& 9.80 & 3.32 
\\ 
\small{AnnexML$\dagger$} 
&  50.00 &  1.56 
\\ 
\small{DiSMEC$\dagger$} 
& 3.75 & 56.02 
\\ 
\small{Parabel$\dagger$} 
& 2.41 & 0.41 
\\ 
\small{PPD-Sparse$\dagger$} 
& 6.00 & 1.71 
\\ 
\small{XML-CNN$\blacklozenge$} 
& 1.49 & 52.23 
\\ 
\small{DeepXML$\blacklozenge$} 
& -- & 18.62 
\\ 
\small{X-Transformer$\ddagger$} 
& 4.20 & 8.22 
\\ 
\small{APLC-XLNet$\blacklozenge$} 
&  1.1  & -- 
\\ 
\small{AttentionXML$\blacklozenge$} 
& 16.56 & 78.30 
\\ 
\small{Astec$\ddagger$} 
& 18.79 & 7.32 
\\ 
\hline 
& & \\ 
\multicolumn{3}{c}{{\bf Amazon-3M}} 
\\ 
\hline 
\small{Method} 
& \small{Model Size (GB)} & \small{Training time (hr)} 
\\ 
\hline 
\small{OGEEC (ours)} 
& 2.50 & 0.002 
\\ 
\small{PfastreXML$\dagger$} 
& 36.79 & 15.74 
\\ 
\small{DiSMEC$\dagger$} 
& 39.71 & $\sim$4995 
\\ 
\small{Parabel$\dagger$} 
& 65.95 & 10.37 
\\ 
\hline 
\end{tabular} 
\end{center} 
\end{table}

\section{Summary and Conclusion} 
\label{sec:Conclusion} 

It is well acknowledged that XML is an open and 
challenging problem, and as a result, the 
existing methods have suggested requirements 
of elaborate modelling and training efforts. 
In this paper, 
we have presented 
a simple 
approach OGEEC for this task 
which achieves quite competitive results and even 
outperforms some of the more complex 
and resource-intensive 
XML techniques on benchmark datasets. 
Further, the empirical promise of OGEEC becomes more 
evident as the dataset size grows, and on 
propensity-scored evaluation metrics 
which 
are more reliable for evaluations in large vocabulary 
problems like XML. 
These surprising empirical results along with little resource requirements also make a strong case for examining this 
approach on other large-scale learning problems.

\section*{Acknowledgments}
YV would like to thank the Department of Science and Technology (India) for the INSPIRE Faculty Award.

\bibliographystyle{IEEEtran} 
\bibliography{egbib} 

\vfill

\end{document}